\ifx\pdfminorversion\undefined\else
\fi
\documentclass[11pt]{article}

\usepackage{PRIMEarxiv}

\usepackage{times}
\usepackage{latexsym}

\usepackage[T1]{fontenc}

\usepackage[utf8]{inputenc}

\usepackage{microtype}

\usepackage{inconsolata}

\usepackage{graphicx}
\usepackage{multicol}
\usepackage[numbers]{natbib}

\usepackage{amsmath}
\usepackage{amssymb}
\usepackage{amsthm}
\usepackage{array}
\usepackage{booktabs}
\usepackage{enumitem}
\usepackage{multirow}
\usepackage{url}
\usepackage{xcolor}
\usepackage{colortbl}
\usepackage{float}
\IfFileExists{algorithm.sty}{
  \usepackage{algorithm}
}{
  \newfloat{algorithm}{tbp}{loa}
  \floatname{algorithm}{Algorithm}
}
\floatstyle{ruled}
\restylefloat{algorithm}
\IfFileExists{algpseudocode.sty}{
  \usepackage{algpseudocode}
}{
  \newcounter{ALGline}
  \newenvironment{algorithmic}[1][]{
    \setcounter{ALGline}{0}
    \begin{list}{}{
      \setlength{\leftmargin}{2.2em}
      \setlength{\labelwidth}{1.8em}
      \setlength{\labelsep}{0.4em}
      \setlength{\itemsep}{0.15ex}
      \setlength{\topsep}{0.25ex}
      \setlength{\parsep}{0pt}
      \setlength{\partopsep}{0pt}
    }
  }{
    \end{list}
  }
  \newcommand{\State}{\stepcounter{ALGline}\item[\arabic{ALGline}.]}
  \newcommand{\Require}{\State \textbf{Require:} }
  \newcommand{\Ensure}{\State \textbf{Ensure:} }
  \newcommand{\Return}{\textbf{return} }
  \newcommand{\Comment}[1]{\hfill\textit{// ##1}}
  \newcommand{\ALGbeginblock}{
    \begin{list}{}{
      \setlength{\leftmargin}{1.6em}
      \setlength{\labelwidth}{0pt}
      \setlength{\labelsep}{0pt}
      \setlength{\itemsep}{0.1ex}
      \setlength{\topsep}{0.1ex}
      \setlength{\parsep}{0pt}
      \setlength{\partopsep}{0pt}
    }
  }
  \newcommand{\ALGendblock}{\end{list}}
  \newcommand{\For}[1]{\State \textbf{for} ##1 \textbf{do}\ALGbeginblock}
  \newcommand{\EndFor}{\ALGendblock}
  \newcommand{\If}[1]{\State \textbf{if} ##1 \textbf{then}\ALGbeginblock}
  \newcommand{\Else}{\ALGendblock\State \textbf{else}\ALGbeginblock}
  \newcommand{\ElsIf}[1]{\ALGendblock\State \textbf{else if} ##1 \textbf{then}\ALGbeginblock}
  \newcommand{\EndIf}{\ALGendblock}
}

\definecolor{darkblue}{rgb}{0, 0, 0.5}
\definecolor{paperteal}{RGB}{48,121,137}
\definecolor{paperteallight}{RGB}{235,244,246}
\definecolor{papergraylight}{RGB}{244,244,244}

\newcommand{\method}{\textsc{ProactAgent}}

\newcommand{\memsys}{\textsc{Experience Base}}
\newcommand{\promethod}{\textsc{ProactRL}}
\newcommand{\evomethod}{\textsc{ExpOnEvo}}
\newcommand{\repo}{\mathcal{D}}

\title{Ask Only When Needed: Proactive Retrieval from Memory and Skills for Experience-Driven Lifelong Agents}

\author{
  Yuxuan Cai$^1$,Wei Li$^{2}$, Jie Zhou$^{1,2}$, Qin Chen$^1$, Xin Li$^{2}$, Bo Zhang$^{2}$, Liang He$^1$ \\
  $^1$ School of Computer Science and Technology, East China Normal University, Shanghai\\ 
  $^2$ Shanghai AI Laboratory \\
  \texttt{\{jzhou, qchen, lhe\}@cs.ecnu.edu.cn}
}

\date{}

\begin{document}

\maketitle

\begin{abstract}
Online lifelong learning agents must decide not only how to act but also when to consult prior experience to continually improve on long-horizon tasks. Existing methods typically retrieve memories passively, such as at task initialization or after each step, and therefore miss knowledge gaps that arise during interaction. We propose \textsc{ProactAgent}, an experience-driven lifelong learning framework for proactive retrieval over a structured Experience Base. \textsc{ProactAgent} continually improves through \textsc{ExpOnEvo}, which jointly updates policies and refines memory, organizing past interactions into factual, episodic, and skill repositories. It further introduces \textsc{ProactRL}, which treats retrieval as an explicit policy action and learns when and what to retrieve. By comparing paired continuations from identical interaction prefixes with and without retrieval, \textsc{ProactRL} provides step-level process rewards that encourage retrieval only when it improves task outcomes or efficiency. Experiments on SciWorld, AlfWorld, and StuLife show that \textsc{ProactAgent} consistently outperforms all baselines, achieving up to 32\% relative improvement in success rate and over 33\% reduction in interaction rounds. Our code will be publicly available at GitHub.
\end{abstract}


\section{Introduction}
Language agents are progressing beyond isolated task-solving toward online lifelong learning, a paradigm in which an agent engages with a continuous stream of interactive tasks while accumulating experience across episodes~\citep{shridhar2020alfworld,wang2022scienceworld,wang2023voyager,cai2025stulife}. Recent advances in chain-of-thought reasoning~\citep{wei2022cot,yao2023tot}, tool use~\citep{schick2023toolformer,yao2023react}, and self-reflection~\citep{madaan2023selfrefine,shinn2023reflexion} have dramatically expanded what an agent can accomplish within a single episode, yet these capabilities yield diminishing returns once the primary bottleneck shifts from within-episode problem-solving to cross-episode online knowledge utilization. Recent benchmarks confirm that even state-of-the-art proprietary models struggle with long-horizon lifelong tasks~\citep{cai2025stulife}, suggesting that raw model capacity alone cannot substitute for the ability to learn from and build upon interaction history.

\begin{figure*}[!t]
    \centering
    \includegraphics[width=0.99\linewidth]{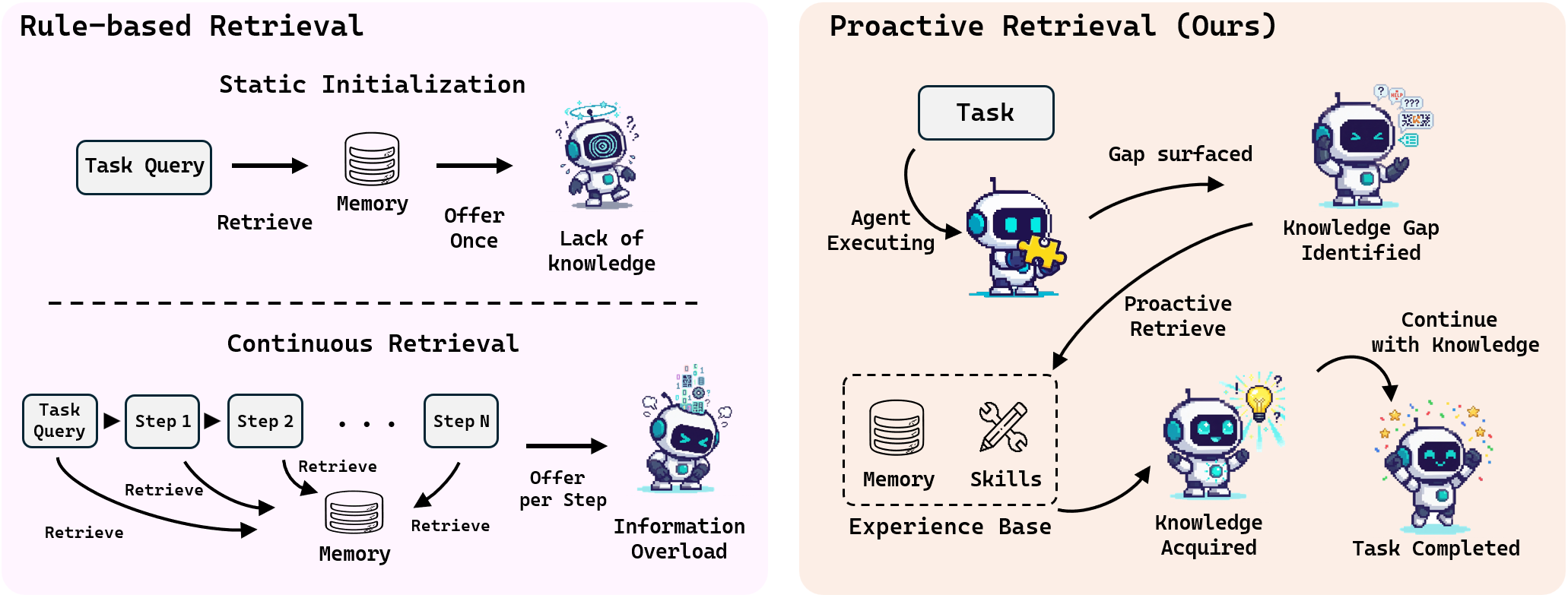}
    \vspace{-2pt}
    \caption{Comparison of retrieval strategies for online lifelong agents. Static initialization provides memory once at task start, failing when task dynamics shift. Continuous retrieval queries at every step, causing context overload. Proactive retrieval (ours) learns when to retrieve through paired-branch process rewards, enabling the agent to recognize knowledge gaps and retrieve precisely when needed.}
    \label{fig:introfig}
     \vspace{-4pt}
\end{figure*}

To realize online experience utilization, a natural and widely adopted approach is to equip agents with an external experience base that stores knowledge accumulated from past interactions~\citep{park2023generative,wang2023voyager,shinn2023reflexion,zhao2024expel,cai2025stulife}. As illustrated in Figure~\ref{fig:introfig}, existing methods mainly differ in how retrieval is triggered. Static initialization~\citep{park2023generative,zhong2023memorybank,wang2023voyager,zhao2024expel} injects memory once at episode onset, allowing agents to start with relevant context but leaving them unable to seek new information when task dynamics shift. Continuous retrieval~\citep{zhang2025memevolve,zhao2024expel,tang2024multihop} performs retrieval at every step, increasing access to external experience but often introducing substantial redundancy and context overload. 
These studies demonstrate the promise of memory-augmented lifelong agents, but also reveal that effective experience utilization depends critically on how retrieval is controlled and how accumulated experience is updated over time.

However, existing lifelong agents still face two fundamental limitations. First, current retrieval mechanisms remain essentially passive: retrieval is triggered by pre-defined positions, externally designed rules, or separate gating modules, rather than being learned as an intrinsic capability of the agent itself. As a result, agents struggle to recognize knowledge gaps during interaction and proactively retrieve the most useful past experience for the current decision. Second, existing online update strategies typically focus on either textual memory accumulation or parameter optimization, while treating the two as largely independent processes. In practice, both are indispensable for lifelong adaptation. Textual memory preserves and expands externalized experience, including facts, episodes, and reusable skills, whereas parameter updates improve the agent's internal policy for future decision making. Updating only one side, or evolving them in isolation, limits the agent's ability to continually improve from interaction history. Therefore, an effective lifelong agent must address both challenges simultaneously: it should learn to retrieve experience proactively, while also jointly evolving its memory and policy online.

To address these challenges, we propose \method, an experience-driven lifelong learning framework that unifies proactive retrieval with experience-enhanced online evolution over a structured experience base. Specifically, \method\ consists of two tightly coupled components. The first component, Experience-Enhanced Online Evolution (\evomethod), jointly improves the agent through memory refinement and policy optimization, enabling it to update both what experience it stores and how it acts during online interaction. The experience base organizes historical interactions into typed repositories, including factual memory, episodic memory, and behavioral skills, so that retrieval can provide both relevant evidence and actionable guidance. The second component, Proactive Reinforcement Learning-based Retrieval (\promethod), formulates retrieval as an explicit policy action and learns both when and what to retrieve through paired-branch process rewards. By comparing continuations from identical interaction prefixes with and without retrieval, \promethod\ provides step-level supervision for retrieval decisions and encourages retrieval only when it leads to better task outcomes or higher efficiency. 

We evaluate \method\ on SciWorld, AlfWorld, and StuLife. It achieves 73.50\% and 71.28\% success rates on SciWorld and AlfWorld with significantly reduced retrieval overhead, while attaining performance competitive with proprietary models on StuLife. These results demonstrate that proactive retrieval, coupled with joint memory and policy evolution, substantially enhances both the effectiveness and efficiency of online lifelong agents.

The main contributions are as follows:
\begin{itemize}[leftmargin=*,itemsep=2pt,topsep=4pt]
\item We propose \method, a proactive online lifelong learning framework enabling agents to continually evolve from interaction history rather than relying on passively invoked experience.

\item We design two key components within \method: Experience-Enhanced Online Evolution (\evomethod), which jointly improves textual memory and policy parameters during online interaction, and Proactive Reinforcement Learning-based Retrieval (\promethod), which learns when and what to retrieve through paired-branch process rewards.

\item We conduct extensive experiments on SciWorld, AlfWorld, and StuLife, showing that \method\ consistently improves lifelong agent performance and retrieval efficiency, and attains performance competitive with proprietary models.
\end{itemize}

\section{Method}
\label{sec:method}

\begin{figure*}[!t]
    \centering
    \includegraphics[width=0.98\linewidth]{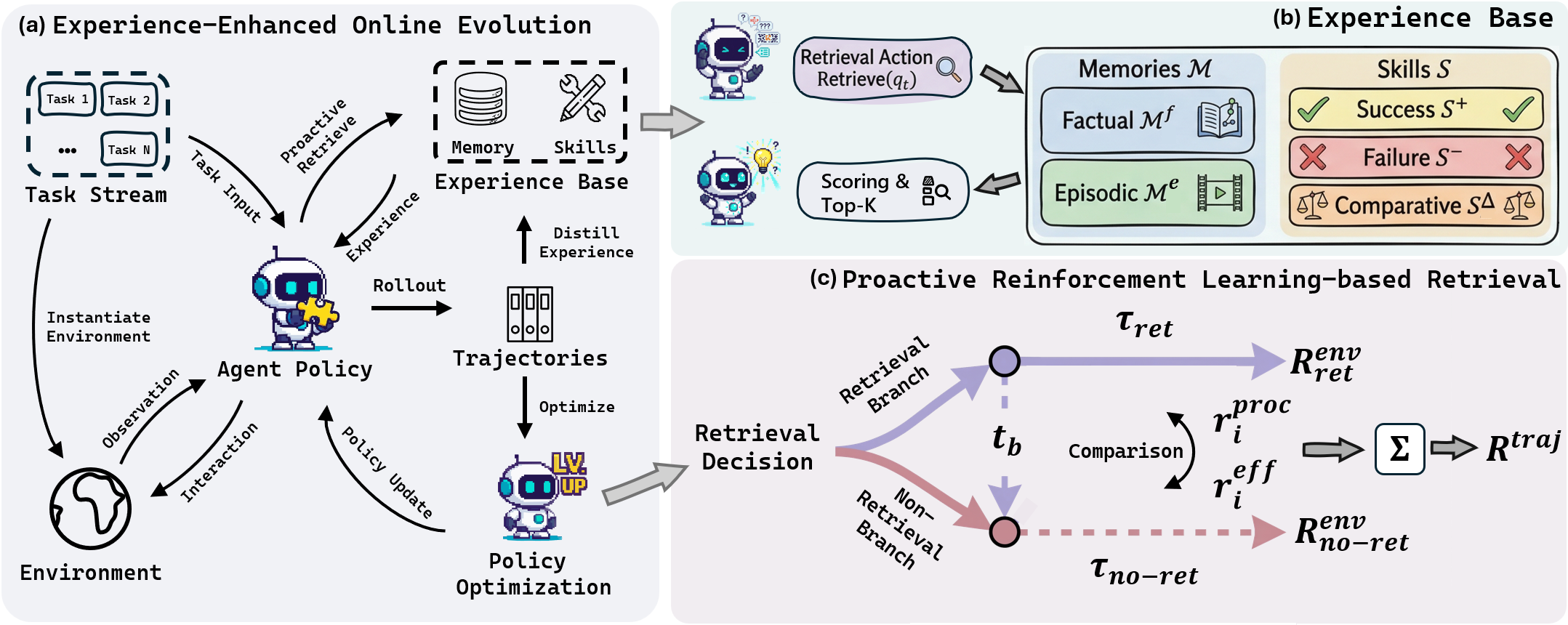}
    \caption{Overview of \method. (\textbf{a})~Experience-Enhanced Online Evolution (\evomethod) closes the loop between acting, experience accumulation, and policy optimization. (\textbf{b})~\memsys\ partitions experience into five typed stores ($\mathcal{M}^\mathrm{f}$, $\mathcal{M}^\mathrm{e}$, $\mathcal{S}^+$, $\mathcal{S}^-$, $\mathcal{S}^\Delta$), so a single query returns complementary evidence and behavioral guidance. (\textbf{c})~Proactive Reinforcement Learning-based Retrieval (\promethod) replays the shared prefix to produce a retrieval branch $\tau^\mathrm{ret}$ and a matched no-retrieval branch $\tau^\mathrm{no\text{-}ret}$; the outcome gap yields a process reward that is positive only when retrieval strictly improves the trajectory, providing supervision for when and what to retrieve.}
    \label{fig:method}
     \vspace{-2pt}
\end{figure*}

    We propose \method, an experience-driven lifelong learning framework enabling agents to continually evolve from interaction history via joint experience refinement, policy optimization, and proactive retrieval over a structured experience base. It comprises two tightly coupled components: Experience-Enhanced Online Evolution (\evomethod) maintains a closed loop of acting, experience accumulation, and policy improvement; and Proactive Reinforcement Learning-based Retrieval (\promethod) formulates retrieval as an explicit policy action, learning when and what to retrieve through step-level supervision.

\subsection{Formal Definition}
We formulate each interactive task instance as a goal-conditioned partially observable Markov decision process. A task instance is defined as
\begin{equation}
x = \langle \mathcal{E}, o_0, g \rangle,
\qquad
\mathcal{E} = (\mathcal{S}, \mathcal{A}, \mathcal{G}, P, R, \Omega, O, \gamma),
\end{equation}
where $\mathcal{S}$, $\mathcal{A}$, $\mathcal{G}$, and $\Omega$ denote the state, action, goal, and observation spaces, respectively; $P(s' \mid s, a)$ represents the transition function; $R(s, a, g)$ represents the goal-conditioned reward; $O(o \mid s', a)$ denotes the observation model; and $\gamma \in [0, 1)$ is the discount factor. Given a horizon $T$, at step $t$, the agent receives a partial observation $o_t \in \Omega$ and constructs the interaction history
\begin{equation}
h_t = (x, o_1, a_1, \ldots, o_t),
\end{equation}
where $a_t$ is the action generated by the model at step $t$. We augment the action space to $\mathcal{A} = \mathcal{A}_{\mathrm{env}} \cup \mathcal{A}_{\mathrm{ret}}$, so the policy can output either an environment action $a_t \in \mathcal{A}_{\mathrm{env}}$ or a retrieval action $a_t = \textsc{Retrieve}(q_t)$, where $q_t$ denotes a natural-language query. If the agent triggers a retrieval action, the returned experience $\mathcal{D}_t$ is appended to the context before the subsequent decision; otherwise, $\mathcal{D}_t = \varnothing$. A trajectory is the ordered sequence $\tau = \bigl((h_t, a_t, \mathcal{D}_t, r_t^{\mathrm{env}})\bigr)_{t=1}^{T}$.

As the agent processes successive tasks, its experience base $\repo$ grows continuously while the behavioral policy adapts through interaction. The central challenge is learning to proactively identify knowledge gaps and selectively retrieve from this expanding repository, since outcome-level rewards alone cannot provide step-level supervision for individual retrieval decisions. \method\ addresses this through two tightly coupled mechanisms (Figure~\ref{fig:method}). Section~\ref{sec:online-evolution} presents Experience-Enhanced Online Evolution, which organizes experience into a structured base and interleaves experience accumulation with policy optimization in a closed loop. Section~\ref{sec:prorl} introduces \promethod, which provides step-level supervision for retrieval decisions through adaptive rollout with paired-branch process rewards, enabling the agent to learn when and what to retrieve. Implementation details are provided in the Appendix~\ref{app:algorithm}.

\subsection{Experience-Enhanced Online Evolution}
\label{sec:online-evolution}

In the online lifelong learning paradigm, an agent engages with a continuous stream of interactive tasks while two resources evolve through successive episodes: the accumulated experience base and the behavioral policy. These resources are naturally complementary. The experience base supplies factual knowledge, episodic precedents, and distilled behavioral patterns from which the policy can learn; a stronger policy, in turn, generates higher-quality trajectories that yield richer entries for the experience base. \method\ closes this loop by unifying experience accumulation and policy optimization into a single iterative cycle (Figure~\ref{fig:method} (a)): given a stream of tasks, the agent (1)~interacts with the environment under the current policy $\pi_\theta$, augmented by proactive retrieval from a structured experience base $\repo$; (2)~distills completed trajectories into typed experience entries that are incorporated into $\repo$; and (3)~updates $\pi_\theta$ via reinforcement learning on the collected trajectories, where retrieval decisions are optimized jointly with task actions through paired-branch process rewards (Section~\ref{sec:prorl}). This creates a self-reinforcing loop: a richer experience base provides more relevant retrieval results, which improves trajectory quality, which in turn yields higher-quality experience entries and stronger policy gradients.

A critical design choice within this framework is how accumulated experience is organized for effective retrieval. At decision time, an interactive agent benefits from two distinct types of support: relevant memory (factual knowledge or similar prior episodes) and behavioral guidance (which action patterns succeed or fail in analogous states). Conflating all experience into a single undifferentiated pool mixes these two modes of support, producing redundant or competing retrieval results. To address this, the experience base maintains a structured repository (Figure~\ref{fig:method} (b)):
\begin{equation}
\begin{aligned}
\repo &= \mathcal{M} \cup \mathcal{S},
\quad
\mathcal{M} = \mathcal{M}^{\mathrm{f}} \cup \mathcal{M}^{\mathrm{e}},\\
\mathcal{S} &= \mathcal{S}^{+} \cup \mathcal{S}^{-} \cup \mathcal{S}^{\Delta},
\end{aligned}
\end{equation}
where $\mathcal{M}^{\mathrm{f}}$ and $\mathcal{M}^{\mathrm{e}}$ denote factual and episodic memories, and $\mathcal{S}^{+}$, $\mathcal{S}^{-}$, $\mathcal{S}^{\Delta}$ denote distilled behavioral skills from successes, failures, and comparative evaluations. Each entry $r \in \repo$ stores textual content, a type label, an embedding vector $e(r)$, and a priority score $p(r)$. Given a query $q_t$, the experience base returns a type-balanced experience:
\begin{equation}
\begin{aligned}
\mathcal{D}_t
&= \bigcup_{C \in \mathcal{C}}
\operatorname{TopK}(q_t, C, k_C),
\quad
\sum_{C \in \mathcal{C}} k_C = K,\\
\mathcal{C}
&= \{\mathcal{M}^{\mathrm{f}}, \mathcal{M}^{\mathrm{e}}, \mathcal{S}^{+},
\mathcal{S}^{-}, \mathcal{S}^{\Delta}\},
\end{aligned}
\end{equation}
where entries within each subset are ranked by:
\begin{equation}
\operatorname{score}(q_t, r) =
\operatorname{sim}(e(q_t), e(r)) + \lambda_p\, p(r).
\end{equation}
The similarity term ensures query relevance, while the priority term softly favors entries with proven utility. This typed decomposition enables a single query to return complementary evidence (what is true, what happened before) and guidance (what to do, what to avoid), mitigating redundancy in monolithic memory systems.

After each episode, the evolution framework distills completed trajectories into new entries that grow the structured base. Factual and episodic entries capture environment states and interaction patterns from individual trajectories. Success and failure skills abstract reusable strategies and error patterns from outcome-specific trajectory subsets. Comparative skills, which encode why one continuation outperforms another, are distilled from the paired rollout branches produced by \promethod\ (Section~\ref{sec:prorl}), exploiting shared prefixes to provide the most localized contrastive signal. Priority scores are updated only for entries that were actually retrieved and associated with improved outcomes, creating a lightweight mechanism that progressively surfaces high-value experience.

Within the online evolution framework, retrieval is not a passive heuristic but an explicit policy action optimized jointly with task behavior.  As the experience base grows through continued interaction, selective retrieval becomes increasingly critical: indiscriminate querying floods context with irrelevant information, while missed opportunities leave the agent repeating past errors. However, standard outcome-level rewards cannot provide the step-level supervision needed to learn which retrieval decisions are beneficial. The next section introduces \promethod, which addresses this through paired-branch process rewards.

\subsection{Proactive Reinforcement Learning-based Retrieval}
\label{sec:prorl}

Within the online evolution framework, the agent must learn when retrieval actively improves outcomes versus when current knowledge suffices. Let $z_t = \mathbf{1}[a_t = \textsc{Retrieve}(q_t)]$ denote whether a retrieval action is executed at step $t$. The core difficulty is that outcome-level rewards cannot isolate whether a specific retrieval decision was beneficial. Upon failure, it remains unclear if retrieval was unnecessary, mistimed, or vague. Therefore, learning proactive retrieval requires step-level supervision answering: did this retrieval improve the trajectory, or would the agent have succeeded without it?

To address this challenge, we introduce an adaptive rollout mechanism that generates comparative evidence (Figure~\ref{fig:method} (c)). Whenever an initial rollout triggers a retrieval action at step $t_b$, we adaptively sample alternative continuations from the same interaction prefix $h_{t_b}$:
\begin{equation}
\begin{aligned}
\tau^{\mathrm{ret}}
&= \tau_{< t_b} \oplus \tau_{\ge t_b}^{\mathrm{ret}},\\
\tau^{\mathrm{no\text{-}ret}}
&= \operatorname{Replay}(\tau_{< t_b})
\oplus \tau_{\ge t_b}^{\mathrm{no\text{-}ret}}.
\end{aligned}
\end{equation}
Here $\tau_{< t_b}$ and $\tau_{\ge t_b}$ denote the ordered prefix and suffix subsequences of $\tau$, respectively, and $\oplus$ denotes sequence concatenation. The rollout $\tau^{\mathrm{ret}}$ preserves the retrieval decision and its subsequent context, whereas $\tau^{\mathrm{no\text{-}ret}}$ explores alternative actions from the same prefix by temporarily suppressing the retrieval action at $t_b$. Because both rollouts share the same history before the branching point, the divergence between them provides a comparative signal: if $\tau^{\mathrm{ret}}$ outperforms $\tau^{\mathrm{no\text{-}ret}}$, the retrieval was proactive and necessary; if not, the retrieval was passive or redundant. This paired-branch comparison directly isolates the utility of the specific retrieval decision, enabling step-level process rewards for proactive control.

For a branched trajectory pair indexed by $i$, let $j(i)$ denote the matched no-retrieval counterpart of rollout $i$, and let $z_i \equiv z_{t_b}^{(i)} \in \{0,1\}$ denote the retrieval indicator at the branching step. \promethod\ defines the trajectory-level reward for reinforcement learning as follows:
\begin{equation}
R_i^{\mathrm{traj}} = R_i^{\mathrm{env}} + r_i^{\mathrm{proc}} + r_i^{\mathrm{eff}},
\end{equation}
where $R_i^{\mathrm{env}} = \sum_{t=1}^{T_i} r_t^{\mathrm{env}}$ denotes the cumulative environment reward of trajectory $i$. For a trajectory $i$ that includes a retrieval action, we evaluate its advantage over the adaptively sampled non-retrieval counterpart $j(i)$ by computing the rollout margin:
\begin{equation}
\Delta_i =
\left(R_i^{\mathrm{env}} - R_{j(i)}^{\mathrm{env}}\right)
+ \lambda_T \frac{T_{j(i)} - T_i}{\max(T_{j(i)}, 1)},
\end{equation}
where $T_i$ denotes the number of interaction steps. The retrieval process reward is then defined as
\begin{equation}
r_i^{\mathrm{proc}} =
\begin{cases}
\alpha, & z_i > 0 \text{ and } \Delta_i > 0, \\
-\alpha, & z_i > 0 \text{ and } \Delta_i < 0, \\
0, & \text{otherwise}.
\end{cases}
\end{equation}
This process reward directly enables proactive retrieval control. It penalizes passive or vague queries ($\Delta_i < 0$) where retrieval provides no benefit over proceeding without it, thereby teaching the agent when not to retrieve and what to avoid querying. Conversely, it provides explicit positive reinforcement ($\alpha$) only when the specific query content actively unlocks a superior or more efficient continuation, teaching the agent exactly when a query is necessary and what information is most effective to retrieve. By learning from these paired comparisons, the agent develops genuine proactivity, i.e., the ability to recognize when current knowledge suffices versus when retrieval is needed.

To further penalize wasteful retrieval actions, we introduce the following efficiency penalty term:
\begin{equation}
\begin{aligned}
r_i^{\mathrm{eff}}
&= - w_q \, \mathbf{1}[\operatorname{repeat}_i]\\
&\quad + \operatorname{clip}\!\Bigl(
w_t \frac{\bar{T}_{g(i)} - T_i}{\max(\bar{T}_{g(i)}, 1)},
-|w_t|, |w_t| \Bigr),
\end{aligned}
\end{equation}
where $\operatorname{repeat}_i$ equals $1$ if rollout $i$ repeats a query string that already appeared earlier in the same trajectory and $0$ otherwise, $g(i)$ denotes the goal associated with rollout $i$, and $\bar{T}_{g(i)}$ is the empirical average length of successful trajectories for goal $g(i)$. This term discourages repeating identical queries and provides rewards for shorter, successful trajectories. The formal proof showing that this process reward provides supervision for proactive retrieval is provided in the appendix. Given a task instance $x$, we sample a group of $G$ rollouts $\{\tau^{(1)}, \ldots, \tau^{(G)}\}$ and compute the standard normalized advantage for GRPO as follows:
\begin{equation}
A_i = \frac{R_i^{\mathrm{traj}} - \operatorname{mean}(\{R_j^{\mathrm{traj}}\}_{j=1}^{G})}{\operatorname{std}(\{R_j^{\mathrm{traj}}\}_{j=1}^{G}) + \epsilon}.
\end{equation}
The optimization of the policy is performed using the standard GRPO objective.

Together, the \evomethod\ and \promethod\ form a self-reinforcing cycle: improved trajectories yield higher-quality experience entries, and this richer experience enables more precise retrieval control via paired-branch comparisons. This mutual reinforcement distinguishes \method\ from isolated approaches, empowering it to proactively identify knowledge gaps and initiate searches, achieving higher task success rates and efficiency.

\section{Experiments}

Our experiments are designed to answer three research questions:
\begin{itemize}[leftmargin=1em, itemsep=2pt, topsep=2pt]
\item \textbf{RQ\,1}: Does learned proactive retrieval outperform passive retrieval baselines? (\S\ref{sec:main-results}, \S\ref{sec:additional-analysis}) 
\item \textbf{RQ\,2}: Does joint co-evolution of experience and policy outperform approaches that evolve either resource in isolation, and what is the contribution of each component? (\S\ref{sec:main-results}, \S\ref{sec:ablation}) 
\item \textbf{RQ\,3}: Does proactive retrieval improve both efficiency and accuracy, and generalize across model scales? (\S\ref{sec:additional-analysis})
\end{itemize}

\subsection{Experimental setup}

\paragraph{Datasets.} We evaluate \method\ on three interactive benchmarks: SciWorld~\citep{wang2022scienceworld}, AlfWorld~\citep{shridhar2020alfworld}, and StuLife~\citep{cai2025stulife}. SciWorld and AlfWorld report the success rate (SR) and the average number of interaction rounds, whereas StuLife reports the SR and StuGPA. For SciWorld and AlfWorld, we adopt the training splits of AgentGym-RL~\citep{xi2025agentgymrl} and evaluate on the test set of AgentGym-RL (SciWorld) and the official test set (AlfWorld). Because StuLife provides no training set, evaluation on this benchmark is restricted to methods capable of evolving entirely online. For methods that involve reinforcement learning, a cold-start phase on the training set precedes policy optimization; methods without a training set (StuLife) rely solely on online evolution.

\paragraph{Baselines.} We evaluate against offline methods that remain static after training (ReAct~\citep{yao2023react}, SFT, GRPO~\citep{shao2024deepseekmath}); online methods that evolve memory (AWM~\citep{wang2024agent}, Reflexion~\citep{shinn2023reflexion}, MemoryBank~\citep{zhong2023memorybank}, Mem0~\citep{chhikara2025mem0}); online Test-time RL~\citep{zuo2025ttrl} that evolves policy parameters via online GRPO without structured memory, and GRPO+Reflexion that combines parameter updates with memory accumulation but relies on passive retrieval.

\paragraph{Evaluation protocol.} Unless noted otherwise, all primary experiments use Qwen2.5-7B-Instruct as the base model; a scaling study with Qwen2.5-3B-Instruct is also reported. \method\ performs online co-evolution in which experience accumulation and parameter adaptation are tightly interleaved at single-instance granularity. Implementation details are provided in Appendix~\ref{app:training}.

\subsection{Main Results}
\label{sec:main-results}

\begin{table*}[t]
\small
\centering
\caption{Main results across three lifelong agent benchmarks. The best results are highlighted in bold.}
\label{tab:main-results}
\begin{tabular}{lccccccc}
\toprule
& \multicolumn{2}{c}{SciWorld} & \multicolumn{2}{c}{AlfWorld} & \multicolumn{2}{c}{StuLife} & \multicolumn{1}{l}{Avg} \\
\cmidrule(lr){2-3} \cmidrule(lr){4-5} \cmidrule(lr){6-7}
Method & SR $\uparrow$ & Rounds $\downarrow$ & SR $\uparrow$ & Rounds $\downarrow$ & SR $\uparrow$ & StuGPA $\uparrow$ & \multicolumn{1}{l}{SR $\uparrow$} \\
\midrule
\multicolumn{8}{l}{\textit{Offline baselines}} \\
Baseline & 2.00 & 33.40 & 27.69 & 32.85 & 2.56 & 6.33 & 10.75 \\
SFT & 19.50 & 30.26 & 54.36 & 26.78 & 5.11 & 7.27 & 26.32 \\
GRPO~\citep{shao2024deepseekmath} & 52.00 & 26.51 & 67.69 & 17.84 & -- & -- & -- \\
\midrule
\multicolumn{8}{l}{\textit{Online baselines}} \\
Test-time RL~\citep{zuo2025ttrl} & 46.50 & 27.87 & 68.71 & 13.99 & 6.71 & 10.28 & 40.64 \\
AWM~\citep{wang2024agent} & 1.00 & 35.65 & 31.28 & 33.56 & 3.51 & 7.60 & 11.93 \\
Reflexion~\citep{shinn2023reflexion} & 4.00 & 33.73 & 36.92 & 38.57 & 6.82 & 11.31 & 15.91 \\
MemoryBank~\citep{zhong2023memorybank} & 2.50 & 30.24 & 32.31 & 37.14 & 6.05 & 7.69 & 13.62 \\
Mem0~\citep{chhikara2025mem0} & 3.00 & 34.94 & 33.85 & 32.11 & 7.34 & 10.83 & 14.73 \\
GRPO+Reflexion & 55.50 & 27.52 & 67.18 & 16.42 & 9.37 & 13.51 & 44.02 \\
\midrule
\multicolumn{8}{l}{\textit{Ours}} \\
\textbf{\method} & \textbf{73.50} & \textbf{18.38} & \textbf{71.28} & \textbf{12.73} & \textbf{12.35} & \textbf{19.26} & \textbf{52.38} \\
\bottomrule
\end{tabular}

\end{table*}

Table~\ref{tab:main-results} presents the main results across three benchmarks. We organize the key findings around the three research questions.

\textbf{Learned proactive retrieval outperforms passive strategies.} Online baselines with experience (AWM, MemoryBank, Mem0) lack learned retrieval control and yield only limited gains over the base model. Even GRPO+Reflexion leaves substantial room for improvement. \method\ achieves an 18.0-point gain over GRPO+Reflexion on SciWorld by learning retrieval as an explicit policy action through paired-branch process rewards.

\textbf{Joint evolution outperforms isolated approaches.} Test-time RL underperforms offline GRPO on SciWorld, suggesting that online updates without structured experience can be unstable on noisy trajectories. Memory-only baselines achieve at most 7.34\% SR on StuLife. \method\ surpasses GRPO+Reflexion and approaches proprietary model performance on StuLife.

\textbf{Proactive control yields efficiency gains alongside accuracy.} \method\ reduces interaction rounds by 33.2\% on SciWorld and 22.5\% on AlfWorld relative to GRPO+Reflexion. By learning when current knowledge suffices and when retrieval is needed, proactive control avoids both unnecessary queries that flood the context and missed retrieval opportunities that force the agent to rediscover known solutions.

\subsection{Ablation Study}
\label{sec:ablation}

\begin{table}[t]
\small
\centering
\caption{Ablation study on SciWorld. 
}
\label{tab:evolution-ablation}
\begin{tabular}{lcc}
\toprule
Variant & SR $\uparrow$ & Rounds $\downarrow$ \\
\midrule
\multicolumn{3}{l}{\textit{Progressive ablation}} \\
\textbf{\method} & \textbf{73.50} & \textbf{18.38} \\
\quad w/o \evomethod\  & 70.50 & 17.15 \\
\quad w/o \promethod\  & 26.50 & 27.80 \\
\quad w/o \evomethod\ + \promethod\  & 5.50 & 30.18 \\
\midrule
\multicolumn{3}{l}{\textit{Component ablation (from w/o online param.\ evol.)}} \\
\quad Replace $\repo$ with Reflexion & 62.50 & 23.15 \\
\quad Remove paired-branch reward & 65.00 & 27.31 \\
\quad Remove cold-start stage & 59.50 & 29.04 \\
\bottomrule
\end{tabular}
\end{table}

\begin{figure*}[t]
\centering
    \includegraphics[width=0.83\linewidth]{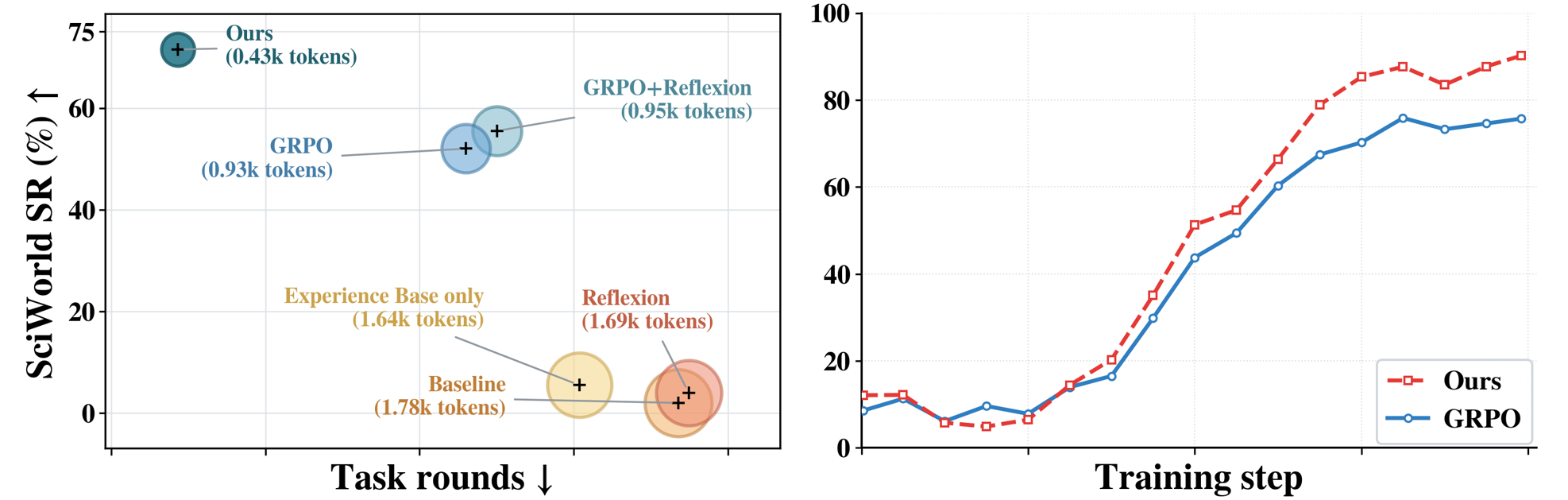}
    \vspace{-2pt}
\caption{Inference efficiency and training dynamics on SciWorld. Left: \method\ achieves higher success rates with fewer interaction rounds and lower token consumption than all baselines, where bubble area indicates average prompt tokens per episode. Right: \method\ consistently outperforms GRPO throughout training, converging to a substantially higher final accuracy.}
\label{fig:efficiency-training}
\end{figure*}

\textbf{Progressive ablation.} We cumulatively remove components from the full \method\ to measure their layered contribution. Removing \evomethod\ results in the offline variant; the full model improves over it by 4.3\% relative in SR, indicating that online parameter evolution is essential to unlock further performance improvements. Further removing \promethod\ reduces SR by 92.2\% relative to the offline variant, confirming that learned proactive retrieval is the most impactful component. Finally, removing \evomethod\ and \promethod\ leaves only the raw $\memsys$ without training or proactive retrieval, demonstrating that structured experience alone is insufficient without online evolution.

\textbf{Component ablation.} Starting from the variant without online parameter evolution, we isolate individual design choices. Replacing the structured $\repo$ with Reflexion reduces SR by 11.3\% relative, confirming that typed decomposition outperforms monolithic memory. Removing paired-branch process rewards while retaining standard GRPO training yields a 7.8\% decrease, demonstrating that step-level supervision improves retrieval learning beyond what outcome-level rewards alone provide. Removing the cold-start phase causes the largest single-component decrease of 15.6\%, confirming that proactive retrieval tool-calling requires supervised initialization before reinforcement learning can effectively shape retrieval decisions.
\begin{table}[t]
\small
\centering
\caption{Component ablation of \memsys.}
\label{tab:memory-skill-ablation}
\setlength{\tabcolsep}{0.5mm}
\begin{tabular}{lccc}
\toprule
Variant & SciWorld SR $\uparrow$ & AlfWorld SR $\uparrow$ & StuLife SR $\uparrow$ \\
\midrule
Full & \textbf{73.50} & \textbf{71.28} & \textbf{12.35} \\
Skill only & 69.00 & 67.69 & 10.74 \\
Memory only & 67.50 & 66.15 & 11.68 \\
\bottomrule
\end{tabular}
\end{table}
\begin{figure*}[t]
\centering
\includegraphics[width=0.97\textwidth, trim=0 363 0 0, clip]{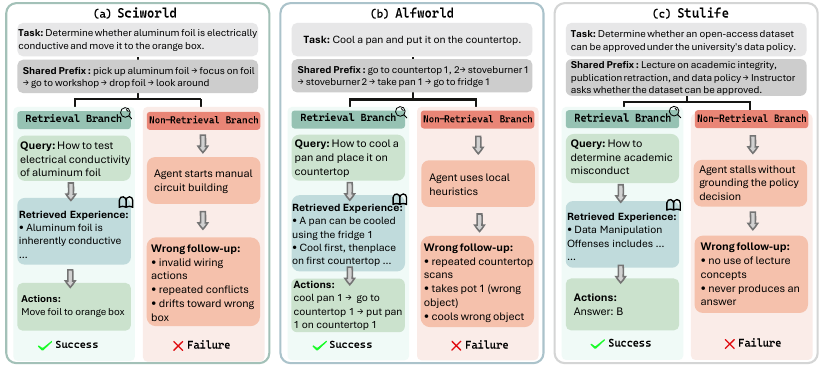}
\vspace{-2pt}
\caption{Case studies across SciWorld, ALFWorld, and StuLife. Each panel contrasts a query branch (green) against a matched no-query branch (red) from the same interaction prefix or task instance. In all three cases, a single targeted retrieval at the action-critical decision point leads to immediate success, while the no-query branch drifts into invalid actions, wrong-object selection, or stalled interaction and fails.}
\label{fig:case-study}
\end{figure*}

\textbf{Experience ablation.} As shown in Table~\ref{tab:memory-skill-ablation}, ablating either degrades performance across all benchmarks, confirming complementary roles. Skill-only performs slightly better on SciWorld and AlfWorld, suggesting action abstractions contribute more in task-oriented environments. Memory-only shows relative strength on StuLife, where long-horizon tasks with cross-task dependencies depend more on factual persistence.

\subsection{Additional Analysis}
\label{sec:additional-analysis}

\textbf{Efficiency analysis.} The left panel of Figure~\ref{fig:efficiency-training} shows that performance gains stem from proactive control rather than extended contexts. \method\ achieves the highest success rate with the fewest interaction rounds and lowest token consumption, reaching 73.50\% SR with only 0.43k tokens per task, compared to GRPO+Reflexion's 0.95k tokens for 55.50\% SR. The right panel's training dynamics show \promethod\ maintaining a consistent advantage over vanilla GRPO from mid-training onward, confirming that proactive retrieval outperforms passive retrieval.

\begin{table}[t]
\centering
\caption{Model scaling on SciWorld with Qwen2.5-3B-Instruct. Our 3B model nearly matches 7B GRPO+Reflexion in SR while using 48\% fewer rounds.}
\label{tab:model-scale}

\begin{tabular}{lcc}
\toprule
Method & SR $\uparrow$ & Rounds $\downarrow$ \\
\midrule
GRPO (3B) & 24.00 & 24.14 \\
GRPO+Reflexion (3B) & 29.50 & 20.09 \\
GRPO+Reflexion (7B) & 55.50 & 27.52 \\
\textbf{\method\ (3B)} & \textbf{53.50} & \textbf{14.35} \\
\bottomrule
\end{tabular}
\end{table}

\textbf{Model scaling.} Table~\ref{tab:model-scale} shows that proactive retrieval remains effective at smaller scales. The 3B model achieves 53.50\% SR (+24.0 points over 3B GRPO+Reflexion), nearly matching 7B GRPO+Reflexion (55.50\%) while requiring 48\% fewer interaction rounds. These results suggest that proactive retrieval can partially offset capacity gaps through more selective decision-making.

\textbf{Benchmark-specific behavior.} Each benchmark reveals distinct benefits of proactive retrieval. SciWorld shows the largest SR and efficiency gains, suggesting retrieval timing is the primary bottleneck in long-horizon scientific tasks. AlfWorld exhibits smaller SR but substantial efficiency improvements, indicating it mainly reduces wasted exploration. On StuLife, \method\ demonstrates strong online adaptation by rivaling leading proprietary systems, suggesting proactive retrieval helps narrow the gap between open-weight and closed-source models in long-dependency tasks.

\textbf{Case study.} Figure~\ref{fig:case-study} contrasts query and no-query branches. Factual retrieval bypasses invalid wiring (SciWorld), procedural retrieval eliminates redundant scanning (ALFWorld), and retrieving misconduct criteria provides knowledge absent from parametric memory (StuLife). In all cases, one targeted query at a decision-critical step replaces a longer, failure-prone trajectory.

\section{Conclusion}

We introduced \method, an experience-driven lifelong learning framework enabling agents to proactively retrieve from a structured experience base and continually improve via interaction. \evomethod\ jointly refines typed memory and policies within a unified co-evolution loop. \promethod\ formulates retrieval as an explicit policy action, learning when and what to retrieve via paired-branch process rewards. Experiments on SciWorld, AlfWorld, and StuLife show \method\ consistently improves task success and interaction efficiency, with a 3B model rivaling 7B passively augmented baselines. These results highlight the effectiveness of proactive retrieval control for long-horizon online lifelong agents.

\section*{Limitations}

\method\ relies on the prefix replayability assumption, which holds in the deterministic simulators used here but may not extend to environments with stochastic dynamics or irreversible state changes. The experience base grows without a capacity bound or learned eviction policy, so unbounded growth could degrade retrieval precision over substantially longer deployment horizons. Experience extraction depends on a separate, larger model, whose errors may propagate into stored entries. All three benchmarks are text-based with discrete actions, and generalization to multi-modal setting remains to be validated.

\section*{Ethical Considerations}

This work focuses on improving the learning efficiency of language model agents in simulated environments (science experiments, household tasks, and campus navigation) and does not involve human subjects, private data, or real-world deployment. All benchmarks are publicly available and contain no personally identifiable information. The experience base is constructed from synthetic agent--environment interactions and stored locally without external dissemination. While our framework enhances autonomous decision-making capabilities, the controlled simulator settings pose minimal risk of societal harm. We acknowledge that deploying lifelong learning agents in open-ended real-world scenarios would require additional safeguards, including human oversight mechanisms and robust content filtering, to prevent unintended or harmful behaviors.

\bibliographystyle{unsrt}
\bibliography{ref}

\newpage

\newpage

\appendix

\section{Related Work}
\label{app:related-work}
\paragraph{Memory-augmented lifelong agents.}
External memory for neural models has progressed from differentiable controllers~\citep{graves2014ntm,weston2015memory,graves2016hybrid,du2025automlgen} through retrieval-augmented generation~\citep{lewis2020rag,borgeaud2022retro} to memory systems designed for interactive agents, where memory must accumulate operational experience across episodes rather than index a static corpus. Recent agent memory systems manage persistent interactions through summarization or hierarchical architectures~\citep{park2023generative,zhong2023memorybank,packer2023memgpt,chhikara2025mem0} and maintain evolving stores that grow alongside continued interaction~\citep{zhang2025memevolve,yang2026autoskill}. As these repositories scale, how retrieval is triggered becomes increasingly critical. Existing approaches differ primarily in triggering mechanism: static initialization~\citep{park2023generative,zhong2023memorybank} injects memory once at episode onset, providing startup context but leaving agents unable to adapt when conditions shift mid-episode; continuous retrieval~\citep{zhang2025memevolve,tang2024multihop} queries at every step, increasing access at the cost of redundancy and context overload; and LLM-gated approaches~\citep{verma2026reflectiverag,zhang2026memskill} employ a separate model to judge necessity, improving flexibility but introducing extra latency. Moreover, most systems store experience in a single undifferentiated repository~\citep{park2023generative,zhong2023memorybank}, conflating factual evidence with behavioral guidance. More critically, retrieval across all three paradigms remains governed by fixed schedules, external rules, or separate modules rather than learned as an intrinsic agent capability, leaving agents unable to recognize when current knowledge is insufficient and proactively seek the most relevant experience.

\paragraph{Retrieval control for LLMs.}
A growing body of work studies how LLMs can control when and what to retrieve. In question answering, FLARE~\citep{jiang2023flare} triggers retrieval based on generation confidence, Self-RAG~\citep{asai2024selfrag} generates reflection tokens to assess retrieval necessity, and Adaptive-RAG~\citep{jeong2024adaptive_rag} routes queries by estimated complexity. In the agent setting, Toolformer and ReAct~\citep{schick2023toolformer,yao2023react} train models to interleave tool calls with reasoning, while IRCoT~\citep{trivedi2023ircot} further interleaves retrieval with chain-of-thought steps. These methods advance retrieval precision but share two limitations in the lifelong agent context. First, they operate over static or slowly changing knowledge bases~\citep{jeong2024adaptive_rag,jiang2023flare}, whereas a lifelong agent must selectively access a repository that grows continuously through its own interaction history. Second, and more critically, retrieval decisions in all these systems remain passive, governed by confidence thresholds, supervised classifiers, or generation-time heuristics that receive no outcome-level feedback on whether a particular retrieval action improved the downstream result~\citep{trivedi2023ircot}. When retrieval introduces noise or displaces useful context, these methods cannot attribute the failure to a specific retrieval decision, because supervision does not isolate individual retrieval actions from the surrounding generation process. 

\paragraph{Online evolution of interactive agents.}
A parallel line of research enables agents to improve autonomously through interaction. Within-episode methods such as iterative self-critique~\citep{madaan2023selfrefine} strengthen single-episode capabilities but do not accumulate cross-episode experience. Cross-episode approaches pursue continual improvement through two largely independent channels. Memory-centric methods accumulate textual experience: Reflexion~\citep{shinn2023reflexion} stores verbal reflections from past failures, Voyager~\citep{wang2023voyager} builds reusable skill libraries, and ExpeL~\citep{zhao2024expel} distills lessons from trial outcomes. Parameter-centric methods optimize the behavioral policy through online reinforcement learning~\citep{xi2025agentgym,zuo2025ttrl}. Recent lifelong benchmarks~\citep{cai2025stulife} further confirm that raw model capacity cannot substitute for sustained online knowledge accumulation. Despite the value of both memory and policy evolution, existing systems treat them in isolation: memory repositories grow without policy-level guidance on retrieval quality, and policy updates proceed without structured access to accumulated experience.

\section{Theoretical Analysis and Proofs}
\label{app:paired-branch-proof}

In this section, we provide a formal analysis of how the adaptive rollout and process reward mechanism in \promethod\ aids credit assignment. Rather than claiming strict variance reduction or strict isolation of causal factors, we show that constructing a matched trajectory pair from a shared interaction prefix yields a localized, counterfactual-correlated empirical signal at the branching step.

To formalize this, write $h_b \equiv h_{t_b}$, $a_b \equiv a_{t_b}$, and $\pi_{\mathrm{env}}(\cdot \mid h_b) \equiv \pi(\cdot \mid h_b, \mathcal{A}_{\mathrm{env}})$. We define the expected environment-level marginal utility of a specific retrieval action $\textsc{Retrieve}(q)$ at history $h_b$ relative to the expected outcome of continuing without retrieval under the current behavior policy:
\begingroup
\small
\begin{equation}
\begin{aligned}
m(h_b, q)
&\triangleq \mu_{\mathrm{ret}}(h_b, q) - \mu_{\mathrm{env}}(h_b),\\
\mu_{\mathrm{ret}}(h_b, q)
&\triangleq
\mathbb{E}_{\tau \sim \pi}
\left[
R^{\mathrm{env}}
\mid h_b, a_b = \textsc{Retrieve}(q)
\right],\\
\mu_{\mathrm{env}}(h_b)
&\triangleq
\mathbb{E}_{\tau \sim \pi}
\left[
R^{\mathrm{env}}
\mid
\substack{
h_b,\\
a_b \sim \pi_{\mathrm{env}}(\cdot \mid h_b)
}
\right].
\end{aligned}
\end{equation}
\endgroup
The paired-branch mechanism does not provide an unbiased estimator of this quantity, but it aims to construct a single-sample empirical proxy that is locally aligned with it. To interpret the pairwise comparison, we require the following assumption regarding the underlying environment and the retrieval system.

\textbf{Assumption 1 }(Prefix Replayability and Consistency)
\label{assum:replay}
For any branching step $t_b$, the environment state and the agent's internal history $h_{t_b}$ can be exactly restored to explore alternative continuations. The transition dynamics $P(s' \mid s, a)$ and observation model $O(o \mid s', a)$ remain consistent across branched rollouts. Furthermore, the memory repository $\repo$ and the retrieval function outputs are deterministic and fixed during these rollouts. For the matched no-retrieval branch, the branching action is sampled from $\pi(\cdot \mid h_{t_b}, \mathcal{A}_{\mathrm{env}})$, and the subsequent suffix is generated from the same behavior policy under the replayed state.

\textbf{Proposition 1 }(Local Credit Assignment via Counterfactual Rollouts)
\label{prop:paired_branch}
Let $h_{t_b}$ be the shared interaction history up to the branching step $t_b$. Consider a matched rollout pair sampled under the \promethod\ objective: $\tau^{\mathrm{ret}}$ (indexed by $i$) with retrieval action $a_{t_b}^{(i)} = \textsc{Retrieve}(q_{t_b})$, and its counterfactual $\tau^{\mathrm{no\text{-}ret}}$ (indexed by $j$) with an environment action $a_{t_b}^{(j)} \in \mathcal{A}_{\mathrm{env}}$. Under Assumption 1, and under the local GRPO approximation in which $\rho \approx 1$ while clipping and the KL term are omitted, write $\pi_i^b \triangleq \pi_\theta(a_{t_b}^{(i)} \mid h_{t_b})$ and $\pi_j^b \triangleq \pi_\theta(a_{t_b}^{(j)} \mid h_{t_b})$. The branching-step contribution to the pairwise policy gradient admits the decomposition
\begin{equation}
\begin{aligned}
g_{t_b}
&\triangleq
A_i \nabla_\theta \log \pi_i^b
+ A_j \nabla_\theta \log \pi_j^b \\
&=
\frac{A_i + A_j}{2}
\nabla_\theta \log \bigl(\pi_i^b \pi_j^b\bigr) \\
&\quad +
\frac{A_i - A_j}{2}
\nabla_\theta \log
\frac{\pi_i^b}{\pi_j^b}.
\end{aligned}
\end{equation}
Hence the local update of the log-odds between the sampled retrieval action and the sampled environment action is governed by the empirical advantage gap $A_i - A_j$. Let $\sigma_R \triangleq \operatorname{std}(\{R_k^{\mathrm{traj}}\}_{k=1}^{G}) + \epsilon$. Moreover, for the matched no-retrieval branch, where $r_j^{\mathrm{proc}} = 0$,
\begin{equation}
\begin{aligned}
A_i - A_j
&= \sigma_R^{-1}\Bigl[
(R_i^{\mathrm{env}} - R_j^{\mathrm{env}})
+ r_i^{\mathrm{proc}}\\
&\qquad\qquad
+ (r_i^{\mathrm{eff}} - r_j^{\mathrm{eff}})
\Bigr],
\end{aligned}
\end{equation}
and therefore
\begin{equation}
\begin{aligned}
A_i - A_j
&= \sigma_R^{-1}\Bigl[
\Delta_i
- \lambda_T \frac{T_j - T_i}{\max(T_j, 1)}
+ r_i^{\mathrm{proc}}\\
&\qquad\qquad
+ (r_i^{\mathrm{eff}} - r_j^{\mathrm{eff}})
\Bigr].
\end{aligned}
\end{equation}
Thus, the branch comparison provides a shaped, noisy empirical proxy that combines the sampled environment margin, the explicit process reward, and the efficiency terms.

We analyze the unclipped surrogate gradient of the GRPO objective for a specific matched pair $(i, j)$ sampled from a group of $G$ rollouts for a task instance $x$. Assuming the policy remains close to the old behavior policy ($\rho \approx 1$) and temporarily omitting the KL divergence penalty for algebraic clarity, the empirical gradient contribution of this pair is:
\begin{equation}
\begin{aligned}
\nabla_\theta \mathcal{J}_{\mathrm{pair}}(\theta)
&\approx
A_i \nabla_\theta \log \pi_\theta(\tau^{(i)} \mid x)\\
&\quad +
A_j \nabla_\theta \log \pi_\theta(\tau^{(j)} \mid x).
\end{aligned}
\end{equation}

Under Assumption 1, both trajectories strictly share the prefix $\tau_{< t_b}$. Let $u_t \triangleq \nabla_\theta \log \pi_\theta(a_t \mid h_t)$ denote the shared-prefix score term and $u_t^{(k)} \triangleq \nabla_\theta \log \pi_\theta(a_t^{(k)} \mid h_t^{(k)})$ denote the branch-specific score term. Decomposing the joint log-probability of each trajectory temporally yields three distinct update phases:
\begin{equation}
\begin{aligned}
\nabla_\theta \mathcal{J}_{\mathrm{pair}}(\theta)
\approx{}&
\sum_{t=1}^{t_b-1}
u_t\bigl(A_i + A_j\bigr)
&& \text{(shared prefix)}\\
&+
A_i u_{t_b}^{(i)}
&& \text{(branch)}\\
&+
A_j u_{t_b}^{(j)}\\
&+
A_i \sum_{t>t_b}
u_t^{(i)}
&& \text{(suffix)}\\
&+
A_j \sum_{t>t_b}
u_t^{(j)}.
\end{aligned}
\label{eq:gradient_grouped}
\end{equation}

Focus on the \textbf{Branching Step Update}. Writing $\pi_i \triangleq \pi_\theta(a_{t_b}^{(i)} \mid h_{t_b})$ and $\pi_j \triangleq \pi_\theta(a_{t_b}^{(j)} \mid h_{t_b})$, we have
\begin{equation}
\begin{aligned}
g_{t_b}
&= A_i \nabla_\theta \log \pi_i + A_j \nabla_\theta \log \pi_j \\
&= \frac{A_i + A_j}{2}
\big( \nabla_\theta \log \pi_i + \nabla_\theta \log \pi_j \big)\\
&\quad
+ \frac{A_i - A_j}{2}
\big( \nabla_\theta \log \pi_i - \nabla_\theta \log \pi_j \big) \\
&= \frac{A_i + A_j}{2}
\nabla_\theta \log (\pi_i \pi_j)
\\
&\quad
+ \frac{A_i - A_j}{2}
\nabla_\theta \log \frac{\pi_i}{\pi_j}.
\end{aligned}
\end{equation}
Therefore, the coefficient multiplying the local log-odds direction $\nabla_\theta \log \frac{\pi_i}{\pi_j}$ is exactly $(A_i - A_j)/2$.

Next, because $A_k$ is defined by group normalization, the sample mean cancels in the pairwise difference:
\begin{equation}
A_i - A_j
= \frac{R_i^{\mathrm{traj}} - R_j^{\mathrm{traj}}}{\sigma_R}.
\end{equation}
For the matched no-retrieval branch, $r_j^{\mathrm{proc}} = 0$, so substituting $R^{\mathrm{traj}} = R^{\mathrm{env}} + r^{\mathrm{proc}} + r^{\mathrm{eff}}$ gives
\begin{equation}
\begin{aligned}
A_i - A_j
&= \sigma_R^{-1}\Bigl[
(R_i^{\mathrm{env}} - R_j^{\mathrm{env}})
+ r_i^{\mathrm{proc}}\\
&\qquad\qquad
+ (r_i^{\mathrm{eff}} - r_j^{\mathrm{eff}})
\Bigr].
\end{aligned}
\end{equation}
Substituting the rollout margin definition $\Delta_i = (R_i^{\mathrm{env}} - R_j^{\mathrm{env}}) + \lambda_T \frac{T_j - T_i}{\max(T_j, 1)}$, we obtain
\begin{equation}
\begin{aligned}
A_i - A_j
&= \sigma_R^{-1}\Bigl[
\Delta_i
- \lambda_T \frac{T_j - T_i}{\max(T_j, 1)}
+ r_i^{\mathrm{proc}}\\
&\qquad\qquad
+ (r_i^{\mathrm{eff}} - r_j^{\mathrm{eff}})
\Bigr].
\end{aligned}
\end{equation}

Under Assumption 1, $R_j^{\mathrm{env}}$ is a single-sample Monte Carlo proxy for the second expectation in $m(h_{t_b}, q_{t_b})$, while $\Delta_i$ further augments this shared-prefix comparison with the length regularizer. Therefore, the branching step anchors the retrieval action against a concrete counterfactual continuation and yields a shaped, outcome-correlated empirical signal. The result is local rather than global: it identifies the exact branch direction affected by the pairwise comparison, but it does not imply that $A_i - A_j$ is an unbiased estimator of $m(h_{t_b}, q_{t_b})$.

\paragraph{Discussion.} 
Proposition 1 shows that the retrieval-versus-no-retrieval preference update at the branching step is controlled by $A_i - A_j$. Under the reward definition of \promethod, this coefficient is determined by the sampled environment margin, the process reward, and the efficiency-difference term, all scaled by the group-normalization factor.

\begin{itemize}
    \item \textbf{Learning when to retrieve.} At the branching step, the pairwise gradient contains the term
    \[
    \frac{A_i - A_j}{2} \nabla_\theta \log \frac{\pi_\theta(\textsc{Retrieve}(q_{t_b}) \mid h_{t_b})}{\pi_\theta(a_{t_b}^{(j)} \mid h_{t_b})}.
    \]
    Therefore, if the retrieval branch yields a larger shaped return than its matched no-retrieval continuation, then $A_i - A_j > 0$ and the local update increases the relative log-probability of retrieving at history $h_{t_b}$. If the retrieval branch underperforms, then $A_i - A_j < 0$ and the update suppresses retrieval at that same decision point. Because the two branches share the identical prefix, this signal is attached to the necessity of retrieval at the current state rather than to unrelated earlier actions.

    \item \textbf{Learning what to retrieve.} The updated action is not a generic retrieval flag, but the instantiated action $\textsc{Retrieve}(q_{t_b})$. Consequently, the same gradient step also reinforces or suppresses the specific query content $q_{t_b}$. Queries that retrieve a useful experience $\mathcal{E}_{t_e}$ increase the downstream environment return, are more likely to induce a positive rollout margin $\Delta_i$, and receive positive process-level reinforcement through $r_i^{\mathrm{proc}}$. In contrast, vague or irrelevant queries tend to produce weak or negative margins and therefore lose relative probability. This is why \promethod\ learns the wording of the query jointly with the timing of retrieval.
\end{itemize}

Consequently, the shared-prefix paired-branch construction explains why \promethod\ can learn both when to retrieve and what to retrieve: the branch comparison supplies a local preference update for triggering retrieval, and that same update is tied to the concrete query instance that produced the observed downstream outcome.

\section{Algorithm Details}
\label{app:algorithm}

This section provides the complete algorithmic description of \method, complementing the formal definitions in Section~\ref{sec:method}. Algorithm~\ref{alg:proact-main} outlines the full online co-evolution loop that interleaves policy optimization with experience accumulation. Algorithm~\ref{alg:paired-branch} details the paired-branch reward computation that supplies step-level supervision for retrieval decisions.

\begin{algorithm}[ht]
\caption{Experience-Enhanced Online Evolution}
\label{alg:proact-main}
\begin{algorithmic}[1]
\Require Base policy $\pi_\theta$, empty experience base $\repo$, task stream $\mathcal{X}$, group size $G$
\Ensure Evolved policy $\pi_\theta$, populated experience base $\repo$
\State \textbf{Cold Start:} Train $\pi_\theta$ via SFT on successful trajectories \Comment{Appendix~\ref{app:pipeline}}
\For{each training iteration}
    \State Sample task batch $\{x_1, \ldots, x_B\}$ from $\mathcal{X}$
    \For{each task $x$ in batch}
        \For{$i = 1, \ldots, G$}
            \State Retrieve initial context $\mathcal{D}_0 \gets \operatorname{TopK}(q_{\mathrm{init}}, \repo, K)$
            \For{$t = 1, \ldots, T$}
                \State Sample action $a_t \sim \pi_\theta(\cdot \mid h_t, \mathcal{D}_{<t})$
                \If{$a_t = \textsc{Retrieve}(q_t)$}
                    \State $\mathcal{D}_t \gets \operatorname{TopK}(q_t, \repo, K)$; append to context
                \Else
                    \State Execute $a_t$ in environment; observe $o_{t+1}, r_t^{\mathrm{env}}$
                \EndIf
            \EndFor
            \State Record trajectory $\tau^{(i)}$
        \EndFor
        \State Construct paired branches via Algorithm~\ref{alg:paired-branch}
        \State Compute $\{R_i^{\mathrm{traj}}\}_{i=1}^{G}$ and group-normalized advantages $\{A_i\}_{i=1}^{G}$
    \EndFor
    \State Update $\pi_\theta$ via GRPO objective $\mathcal{J}_{\promethod}(\theta)$ \Comment{advantages in Eq.~11}
    \State Extract typed entries from completed trajectories; update $\repo$ \Comment{Appendix~\ref{app:extraction}}
\EndFor
\end{algorithmic}
\end{algorithm}

\begin{algorithm}[ht]
\caption{\promethod\ Paired-Branch Reward Computation}
\label{alg:paired-branch}
\begin{algorithmic}[1]
\Require Trajectory $\tau^{(i)}$ with retrieval at steps $\{t_1, \ldots, t_n\}$, policy $\pi_\theta$, experience base $\repo$
\Ensure Trajectory-level reward $R_i^{\mathrm{traj}}$
\If{$n \ge 3$}
    \State Select $t_b$ uniformly from $\{t_2, \ldots, t_{n-1}\}$ \Comment{Prefer interior steps}
\Else
    \State Select $t_b$ uniformly from $\{t_1, \ldots, t_n\}$
\EndIf
\State Restore environment state at $h_{t_b}$ via prefix replay \Comment{Assumption 1}
\State Suppress retrieval at $t_b$; sample $a_{t_b}^{(j)} \sim \pi_\theta(\cdot \mid h_{t_b}, \mathcal{A}_{\mathrm{env}})$
\State Generate no-retrieval continuation $\tau^{(j)}$ from $(h_{t_b}, a_{t_b}^{(j)})$ under $\pi_\theta$
\State Evaluate: $R_i^{\mathrm{env}} \gets \sum_t r_t^{\mathrm{env}}(\tau^{(i)})$, \quad $R_j^{\mathrm{env}} \gets \sum_t r_t^{\mathrm{env}}(\tau^{(j)})$
\State Compute rollout margin: $\Delta_i \gets (R_i^{\mathrm{env}} - R_j^{\mathrm{env}}) + \lambda_T \frac{T_j - T_i}{\max(T_j, 1)}$
\If{$\Delta_i > 0$}
    \State $r_i^{\mathrm{proc}} \gets +\alpha$ \Comment{Retrieval improved outcome}
\ElsIf{$\Delta_i < 0$}
    \State $r_i^{\mathrm{proc}} \gets -\alpha$ \Comment{Retrieval was redundant or harmful}
\Else
    \State $r_i^{\mathrm{proc}} \gets 0$
\EndIf
\State $r_i^{\mathrm{eff}} \gets -w_q \cdot \mathbf{1}[\mathrm{repeat}_i] + \operatorname{clip}\!\left(w_t \frac{\bar{T}_{g(i)} - T_i}{\max(\bar{T}_{g(i)}, 1)},\; -|w_t|,\; |w_t|\right)$
\State \Return $R_i^{\mathrm{traj}} \gets R_i^{\mathrm{env}} + r_i^{\mathrm{proc}} + r_i^{\mathrm{eff}}$
\end{algorithmic}
\end{algorithm}

\paragraph{Interaction protocol.}
During each rollout, the agent processes the task prompt augmented with an initial retrieval context $\mathcal{D}_0$ drawn from $\repo$. At each subsequent step, the policy produces either an environment action or a retrieval action $\textsc{Retrieve}(q_t)$. When a retrieval action is triggered, the experience base returns a type-balanced set of entries $\mathcal{D}_t$ (Equation~4 in Section~\ref{sec:online-evolution}), which is appended to the agent's context before the next decision. The policy treats retrieval as a regular action within the augmented action space $\mathcal{A} = \mathcal{A}_{\mathrm{env}} \cup \mathcal{A}_{\mathrm{ret}}$, requiring no separate gating module or confidence threshold.

\paragraph{Experience update protocol.}
After each training iteration, completed trajectories are processed by the extraction model (Appendix~\ref{app:extraction}) to produce typed entries. Factual and episodic memories are extracted from individual trajectories, success and failure skills are distilled from outcome-specific subsets, and comparative skills are generated from paired branches produced by Algorithm~\ref{alg:paired-branch}. All new entries are deduplicated against the existing repository before insertion. Priority scores of retrieved entries associated with successful trajectories are incremented, progressively surfacing high-utility experience.

\section{Experience Base Implementation}
\label{app:implementation}

This section details the implementation of the structured experience base introduced in Section~\ref{sec:online-evolution}, whose role in the overall system is illustrated in Algorithm~\ref{alg:proact-main}. We describe the repository schema, the retrieval backend, and the entry extraction procedure.

\subsection{Repository Schema}
\label{app:schema}

The \memsys\ maintains five typed entry families, organized into two complementary groups. The memory group ($\mathcal{M}$) provides evidence by recording what is true or what has happened, while the skill group ($\mathcal{S}$) provides behavioral guidance by encoding what to do or what to avoid. Table~\ref{tab:entry-families} summarizes the schema.

\begin{table*}[t]
\centering
\caption{The five entry families in the \memsys. Memory entries ($\mathcal{M}$) supply factual evidence, while skill entries ($\mathcal{S}$) supply behavioral guidance. This separation ensures that a single query can retrieve complementary support from both groups.}
\label{tab:entry-families}
\resizebox{\linewidth}{!}{
\begin{tabular}{llll}
\toprule
Group & Entry Type & Content & Source \\
\midrule
\multirow{2}{*}{Memory ($\mathcal{M}$)}
  & Factual ($\mathcal{M}^{\mathrm{f}}$) & Environment facts, tool outputs, persistent states & Per-trajectory \\
  & Episodic ($\mathcal{M}^{\mathrm{e}}$) & Local plans, constraints, trajectory-specific reminders & Per-trajectory \\
\midrule
\multirow{3}{*}{Skill ($\mathcal{S}$)}
  & Success ($\mathcal{S}^{+}$) & Reusable strategies from successful completions & Successful trajectories \\
  & Failure ($\mathcal{S}^{-}$) & Error patterns and corrective rules & Failed trajectories \\
  & Comparative ($\mathcal{S}^{\Delta}$) & Contrastive insights on why one continuation outperforms another & Paired A/B branches \\
\bottomrule
\end{tabular}}
\end{table*}

Each entry $r \in \repo$ stores four fields: (1) a natural-language \texttt{when\_to\_use} description that specifies the triggering context, (2) a \texttt{content} field containing the actual knowledge or guidance, (3) an embedding vector $e(r)$ computed from the \texttt{when\_to\_use} field, and (4) a priority score $p(r)$ that is updated based on retrieval utility. The \texttt{when\_to\_use} field also serves as the exact-match deduplication key, preventing the repository from accumulating near-identical entries.

\subsection{Retrieval Backend}
\label{app:retrieval-backend}

The retrieval backend is built on the following components:

\begin{itemize}[leftmargin=*,itemsep=2pt,topsep=4pt]
\item \textbf{Encoder.} All \texttt{when\_to\_use} fields and runtime queries are embedded using \texttt{all-MiniLM-L6-v2} sentence embeddings (384 dimensions).

\item \textbf{Vector index.} Embeddings are L2-normalized and stored in a FAISS inner-product index, which is equivalent to cosine similarity search over normalized vectors.

\item \textbf{Scoring.} Given a query $q_t$, each candidate entry $r$ is scored as $\operatorname{score}(q_t, r) = \operatorname{sim}(e(q_t), e(r)) + \lambda_p\, p(r)$, combining semantic similarity with the priority term (Section~\ref{sec:online-evolution}).

\item \textbf{Type-balanced retrieval.} The retrieval budget $K$ is divided equally across the five entry types, with one slot allocated to each by default. This prevents any single type from dominating the returned context and encourages complementary evidence.

\item \textbf{Priority update.} Only entries that are actually retrieved during a trajectory receive priority updates. Specifically, when a trajectory succeeds, the priority of each retrieved entry is incremented by one. Entries that are stored but never retrieved receive no update. This mechanism progressively surfaces high-utility entries without requiring explicit supervision.

\item \textbf{Deduplication.} Exact-match deduplication on the \texttt{when\_to\_use} field is enabled by default. The current implementation does not impose a fixed repository capacity cap or employ a learned eviction policy.
\end{itemize}

\subsection{Entry Extraction and Maintenance}
\label{app:extraction}

After each episode, completed trajectories are filtered to retain episodes with valid response tokens and then grouped by task instance. Within each group, a dedicated extraction model (Qwen3-32B) produces typed entries according to the following procedure:

\begin{enumerate}[leftmargin=*,itemsep=2pt,topsep=4pt]
\item \textbf{Factual and episodic memories} are extracted from individual trajectories via summarization. The extractor produces at most two entries of each type per trajectory, focusing on environment facts and trajectory-specific plans or constraints.

\item \textbf{Success and failure skills} are distilled from outcome-specific trajectory subsets. Each distiller returns one to three JSON-formatted entries that encode reusable strategies (from successes) or corrective rules (from failures).

\item \textbf{Comparative skills} are distilled from matched trajectory pairs. Paired A/B branches produced by \promethod\ are prioritized because they share the same task prefix and therefore expose the most localized contrastive signal. When such pairs are unavailable, the extractor falls back to outcome-ranked trajectory pairs from the same task group.
\end{enumerate}

The complete prompts used for each extraction type are provided in Appendix~\ref{app:prompts}.

\begin{table*}[t]
\centering
\caption{Core hyperparameters of the \method\ implementation. Annealing schedule entries are formatted as (no-retrieval fraction, learning rate warmup ratio).}
\label{tab:appendix-hparams}
\small
\begin{tabular}{>{\raggedright\arraybackslash}p{0.30\linewidth}>{\raggedright\arraybackslash}p{0.56\linewidth}}
\toprule
Component & Value \\
\midrule
Base model & Qwen2.5-7B-Instruct (main), Qwen2.5-3B-Instruct (scaling study) \\
Extraction model & Qwen3-32B (memory and experience extraction) \\
Learning rate& 1e-6\\
Training epoch& 3\\
Batch size& 16\\
GRPO group size&8\\
Training budget & 8 H100 GPUs for approximately 40 hours \\
Retriever embedding model & \texttt{all-MiniLM-L6-v2} \\
Vector index & FAISS inner-product index over normalized embeddings \\
Retrieval size & $k = 5$ \\
Type quota & 1 each for factual / episodic / success / failure / comparative \\
Paired-branch process reward ($\alpha$) & $0.5$ \\
Repeated-query penalty weight ($w_q$) & $0.5$ \\
Step-efficiency weight ($w_t$) & $0.25$ \\
Annealing schedule & calibration: $(0.5, 0.2)$, transition: $(0.25, 0.3)$, refinement: $(0.0, 0.5)$ \\
No-retrieval branch & Retrieval suppressed \\
\bottomrule
\end{tabular}
\end{table*}
\section{Training Protocol}
\label{app:training}

This section describes the full training pipeline, including the cold-start phase, the paired-branch rollout procedure, and the retrieval annealing schedule. These stages correspond to the procedural flow outlined in Algorithm~\ref{alg:proact-main} and Algorithm~\ref{alg:paired-branch}. The hyperparameters used in all experiments are summarized in Table~\ref{tab:appendix-hparams}.

\subsection{Pipeline Overview}
\label{app:pipeline}

The training pipeline proceeds through six sequential stages:

\begin{enumerate}[leftmargin=*,itemsep=2pt,topsep=4pt]
\item \textbf{Cold start.} The base policy is trained via supervised learning on successful trajectories to learn the interaction format, valid action syntax, and retrieval-tag conventions. This stage is critical for initializing the policy with sufficient tool-calling competence before reinforcement learning begins (as confirmed by the ablation in Section~\ref{sec:ablation}).

\item \textbf{Rollout sampling.} Multiple rollouts are sampled for each training prompt under the current policy. A portion of these rollouts is configured as no-retrieval trajectories through the \texttt{retrieval\_enabled} switch, whose probability is annealed across training phases.

\item \textbf{Paired-branch construction.} When paired branching is active, the system identifies retrieval-trigger steps in retrieval-enabled rollouts, replays the corresponding prefixes, and creates matched no-retrieval branches (Section~\ref{app:paired-branch}).

\item \textbf{Reward computation.} The environment outcome is combined with the paired-branch process reward and the efficiency bonus to produce the \promethod\ trajectory-level reward (Section~\ref{sec:prorl}).

\item \textbf{Policy update.} The policy is updated using GRPO-style group normalization with PPO-style clipped surrogate optimization.

\item \textbf{Experience base update.} The experience base $\repo$ is updated by extracting factual, episodic, success, failure, and comparative entries from the new trajectories (Appendix~\ref{app:extraction}).
\end{enumerate}

This organization ensures that policy learning and memory growth remain tightly interleaved throughout training, realizing the co-evolution loop described in Section~\ref{sec:online-evolution}.

\subsection{Paired-Branch Rollout Procedure}
\label{app:paired-branch}

When a rollout $\tau$ contains one or more retrieval actions, the system constructs a matched no-retrieval branch as follows. If $\tau$ contains at least three retrieval actions, the branching step $t_b$ is selected uniformly at random from the interior retrieval steps of $\tau$, excluding the first and last retrieval actions. Boundary retrievals are excluded when possible because they are more susceptible to confounds from initialization effects (at the first step) or termination effects (at the last step), which could introduce noise into the process reward signal. If $\tau$ contains fewer than three retrieval actions, $t_b$ is sampled uniformly from all available retrieval steps. The environment state at $t_b$ is restored via prefix replay (Assumption 1), and the no-retrieval branch $\tau^{\mathrm{no\text{-}ret}}$ is generated by suppressing the retrieval action at $t_b$ and sampling subsequent actions from the current policy.

\subsection{Retrieval Annealing}
\label{app:annealing}

During training, the probability of disabling retrieval is annealed across three phases:

\begin{itemize}[leftmargin=*,itemsep=2pt,topsep=4pt]
\item \textbf{Calibration phase.} Corresponding branch paths were generated for all retrieval trajectories, ensuring abundant no-retrieval trajectories for paired A/B branch construction.

\item \textbf{Transition phase.} The no-retrieval fraction decreases to $25\%$, maintaining some paired branches while allowing the retrieval-enabled policy to receive more optimization signal. 

\item \textbf{Refinement phase.} All rollouts are retrieval-enabled ($0\%$ disabled), concentrating optimization on the fully proactive policy.
\end{itemize}

The detailed hyperparameters are reported in Table~\ref{tab:appendix-hparams}.

\section{Experience Extraction Prompts}
\label{app:prompts}

This section presents the complete prompts used by the extraction model (Qwen3-32B) to distill completed trajectories into typed experience entries for the \memsys\ . Each prompt targets a specific entry family and is designed to produce structured JSON outputs with two fields: \texttt{when\_to\_use} (the triggering context for future retrieval) and \texttt{content} (the actual knowledge or guidance). All prompts enforce domain-agnostic language to ensure that extracted entries generalize across task types rather than overfitting to specific benchmark instances.

\begin{figure*}[!t]
\begin{center}
\includegraphics[width=\textwidth]{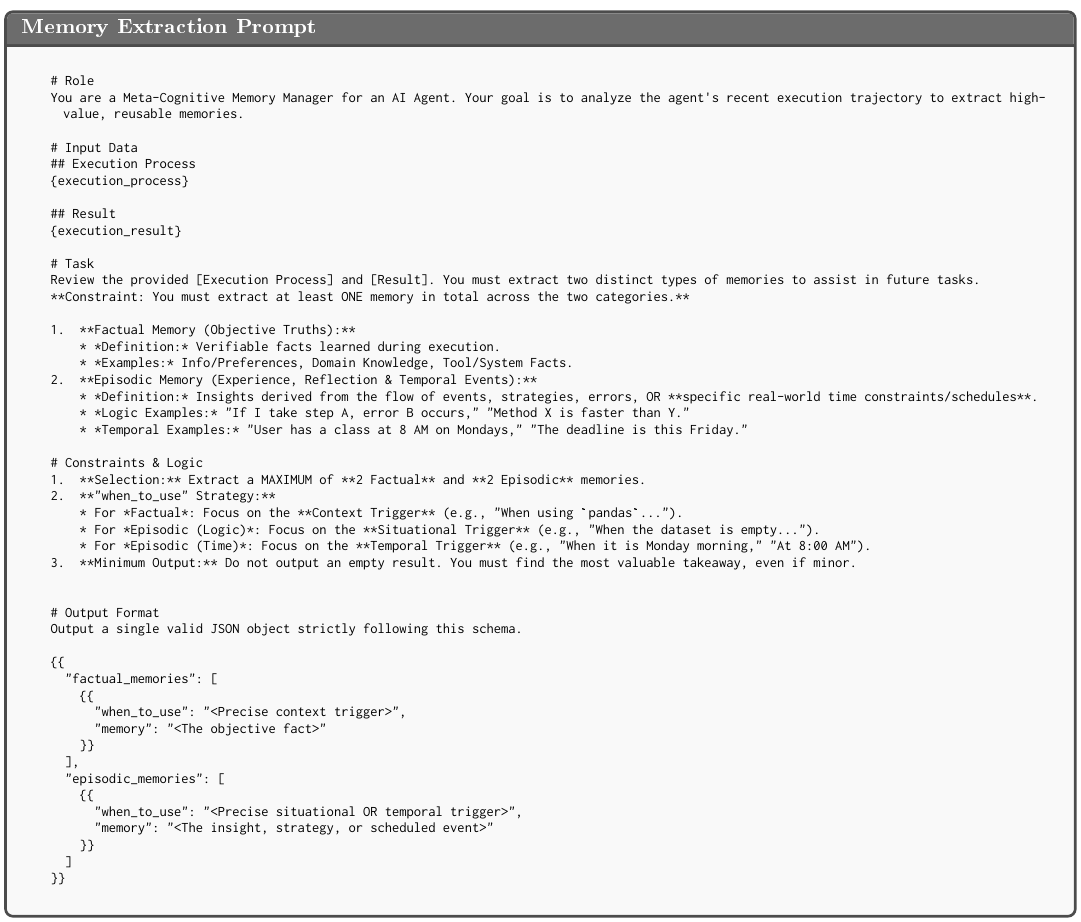}
\end{center}
\end{figure*}
\begin{figure*}[!t]
\begin{center}
\includegraphics[width=\textwidth]{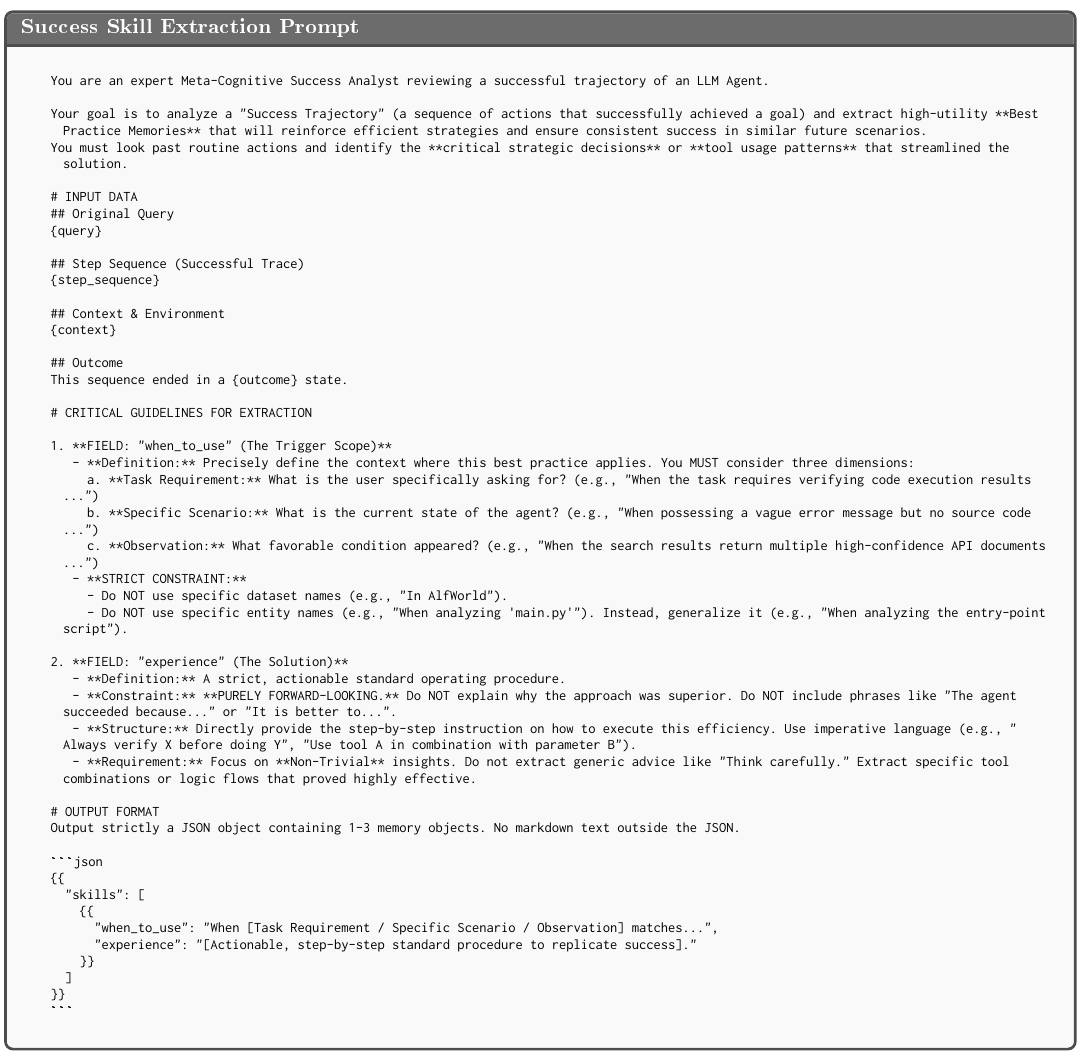}
\end{center}
\end{figure*}
\begin{figure*}[!t]
\begin{center}
\includegraphics[width=\textwidth]{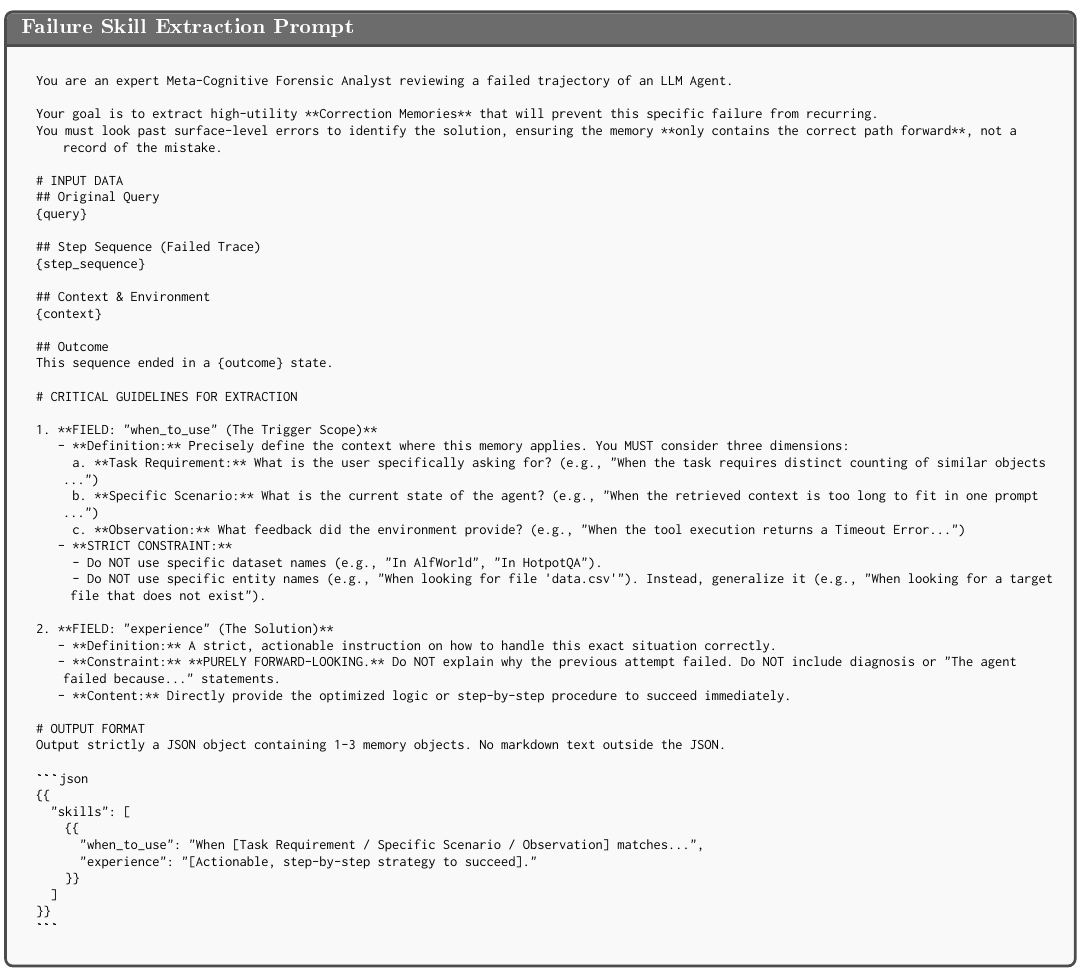}
\end{center}
\end{figure*}
\begin{figure*}[!t]

\begin{center}
\includegraphics[width=\textwidth]{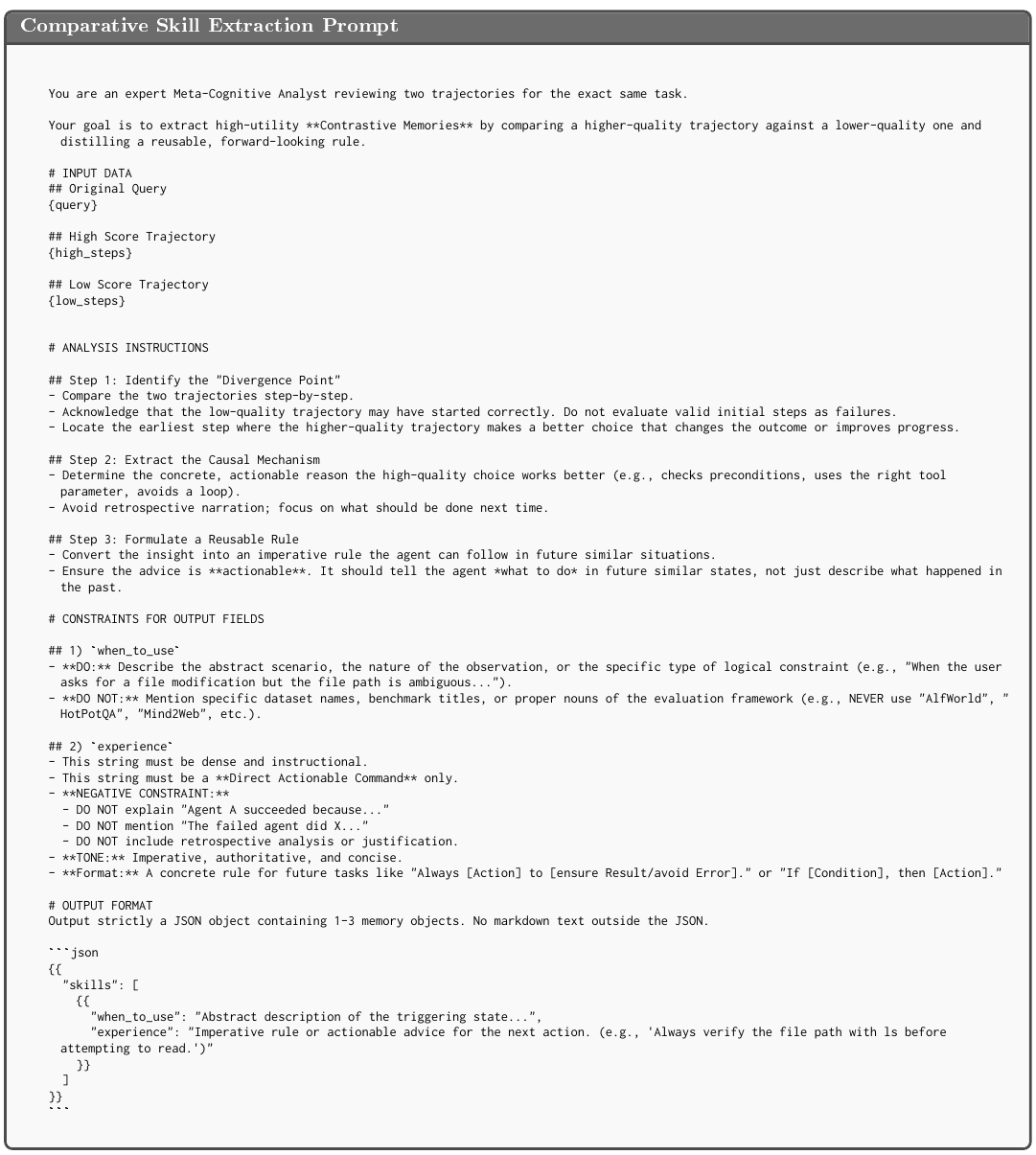}
\end{center}
\end{figure*}

\subsection{Memory Extraction Prompt}
\label{app:prompt-memory}

The following prompt extracts factual and episodic memories from individual trajectories. Factual memories capture verifiable environment facts, while episodic memories capture experiential insights and temporal constraints.

\subsection{Success Skill Extraction Prompt}
\label{app:prompt-success}

The following prompt distills reusable best practices from successful trajectories. It identifies the critical strategic decisions that enabled task completion and formulates them as forward-looking, imperative rules.

\subsection{Failure Skill Extraction Prompt}
\label{app:prompt-failure}

The following prompt extracts corrective rules from failed trajectories. It focuses exclusively on the forward-looking solution rather than the diagnosis of past mistakes, ensuring that extracted entries are immediately actionable.

\subsection{Comparative Skill Extraction Prompt}
\label{app:prompt-comparative}

The following prompt distills contrastive insights from matched trajectory pairs. It compares a higher-quality trajectory against a lower-quality one sharing the same task, identifies the divergence point, and formulates a reusable rule that captures why the better continuation succeeded. Comparative entries produced from \promethod\ paired branches carry the strongest contrastive signal because the two trajectories share the same interaction prefix up to the branching step.

\end{document}